%% file: main.tex
\documentclass[dvipsnames]{article}
\usepackage[preprint]{colm2025_conference}
\vspace{0pt} 
\usepackage{parskip}
\input{math_commands.tex}

\usepackage{hyperref}
\usepackage{url}
\usepackage{booktabs}
\usepackage{multirow}
\usepackage{graphicx}
\usepackage{float}
\usepackage{titlesec}
\usepackage{latexsym}
\usepackage{caption}
\usepackage{colortbl}
\usepackage[most]{tcolorbox}
\usepackage{subcaption}
\usepackage{helvet}
\usepackage{xfrac}
\usepackage{algorithm}
\usepackage{algpseudocode}
\usepackage{wrapfig}
\usepackage{bbm}
\usepackage{amsmath}   
\usepackage{amssymb}   
\usepackage{makecell}
\usepackage{todonotes}
\usepackage{bigfoot}

\usepackage{booktabs}      
\usepackage{multirow}       
\usepackage{arydshln}      
\usepackage{caption}       
\usepackage{subcaption}    
\usepackage{graphicx}      
\usepackage{xcolor}         
\usepackage{colortbl}       
\usepackage{array}          
\usepackage{enumitem}
\usetikzlibrary{calc}
\usepackage{yhmath}

\definecolor{grpo}{HTML}{FAD4D4}   
\definecolor{gfpo}{HTML}{BEE4D0}   
\definecolor{comparison}{HTML}{B3D9FF} 

\definecolor{grayrowcolor}{RGB}{220,220,220} 

\definecolor{commentcolour}{rgb}{0.3,0.7,0.2}
\definecolor{lightblue}{RGB}{245, 250, 250} 
\definecolor{blue}{RGB}{77, 174, 172} 

\definecolor{darkpink}{RGB}{255, 105, 180} 

\newcommand{\best}[1]{\cellcolor{gray!20}\textbf{#1}}

\def\shownotes{0}  
\ifnum\shownotes=1
\newcommand{\authnote}[2]{[#1: #2]}
\else
\newcommand{\authnote}[2]{}
\fi

\newtcolorbox{findings}{
    enhanced,
    breakable,
    colback=lightblue,         %
    colframe=blue,        %
    boxrule=1.5pt,
    arc=0.25em,
    left=1em,
    right=1em,
    top=1em,
    bottom=0.75em,
    before=\vspace{1em},
    overlay unbroken and first={
        \node[
            fill=blue,
            text=white,
            font=\bfseries,
            anchor=west,
            inner xsep=0.75em,
            inner ysep=0.5em,
            rounded corners=0.25em
        ] 
        at ([xshift=0.75em]frame.north west) {Finding};
    }
}

\tcbset{
  aibox/.style={
    width=\textwidth,
    top=0pt, bottom=0pt, left=5pt, right=5pt,
    colback=white,
    colframe=black,
    colbacktitle=black,
    enhanced,
    center,
    attach boxed title to top left={yshift=-0.1in,xshift=0.15in},
    boxed title style={boxrule=0pt,colframe=white,},
  }
}
\newtcolorbox{AIbox}[2][]{aibox,title=#2,#1}
\definecolor{user}{HTML}{F2EFE7}  
\definecolor{model}{HTML}{81E7AF} 

\newcounter{mySubSec}[section] 

\title{The Art of Scaling Test-Time Compute for Large Language Models}

\newcommand{\authorsep}{\hspace{2ex}}
\newcommand{\instsep}{\hspace{2ex}}

 {\author{Aradhye Agarwal$^{m, \omega}$\thanks{Corresponding author: \texttt{aradhye.agarwal@gmail.com}} \authorsep 
Ayan Sengupta$^\omega$ \authorsep Tanmoy Chakraborty$^\omega$\\ 
$^{m}$Microsoft Research \instsep\\
$^{\omega}$Indian Institute of Technology Delhi
}}

\begin{document}

\maketitle

\input{sections/abstract}

\input{sections/introduction}
\input{sections/preliminaries}
\input{sections/prob_diff}

\input{sections/results}
\input{sections/analysis}
\input{sections/recipe}
\input{sections/conclusion}
\bibliography{colm2025_conference}
\bibliographystyle{colm2025_conference}
\newpage
\input{sections/appendix}
\end{document}

%% file: math_commands.tex
\usepackage{amsmath,amsfonts,bm}

\def\eqref#1{equation~\ref{#1}}

\def\1{\bm{1}}

\DeclareMathAlphabet{\mathsfit}{\encodingdefault}{\sfdefault}{m}{sl}
\SetMathAlphabet{\mathsfit}{bold}{\encodingdefault}{\sfdefault}{bx}{n}



%% file: sections/abstract.tex
\begin{abstract}
Test-time scaling (TTS)—the dynamic allocation of compute during inference—is a promising direction for improving reasoning in large language models (LLMs). However, a systematic comparison of well-known TTS strategies under identical conditions is missing, and the influence of model type and problem difficulty on performance remains unclear. To address these gaps, we conduct the first large-scale study of TTS, spanning over thirty billion tokens generated using eight open-source LLMs (7B to 235B parameters), across four reasoning datasets. We observe three consistent trends: (1) no single TTS strategy universally dominates; (2) reasoning models exhibit distinct trace-quality patterns across problem difficulty and trace length, forming short-horizon and long-horizon categories; and (3) for a given model type, the optimal TTS performance scales monotonically with compute budget. Based on these insights, we provide a practical recipe for selecting the best TTS strategy, considering problem difficulty, model type, and compute budget, providing a practical guide to effective inference-time scaling.\footnote{The source code is available at \verb|https://github.com/Aradhye2002/art_of_tts|}
\end{abstract}

%% file: sections/introduction.tex
\begin{figure}[h]
    \includegraphics[width=\textwidth]{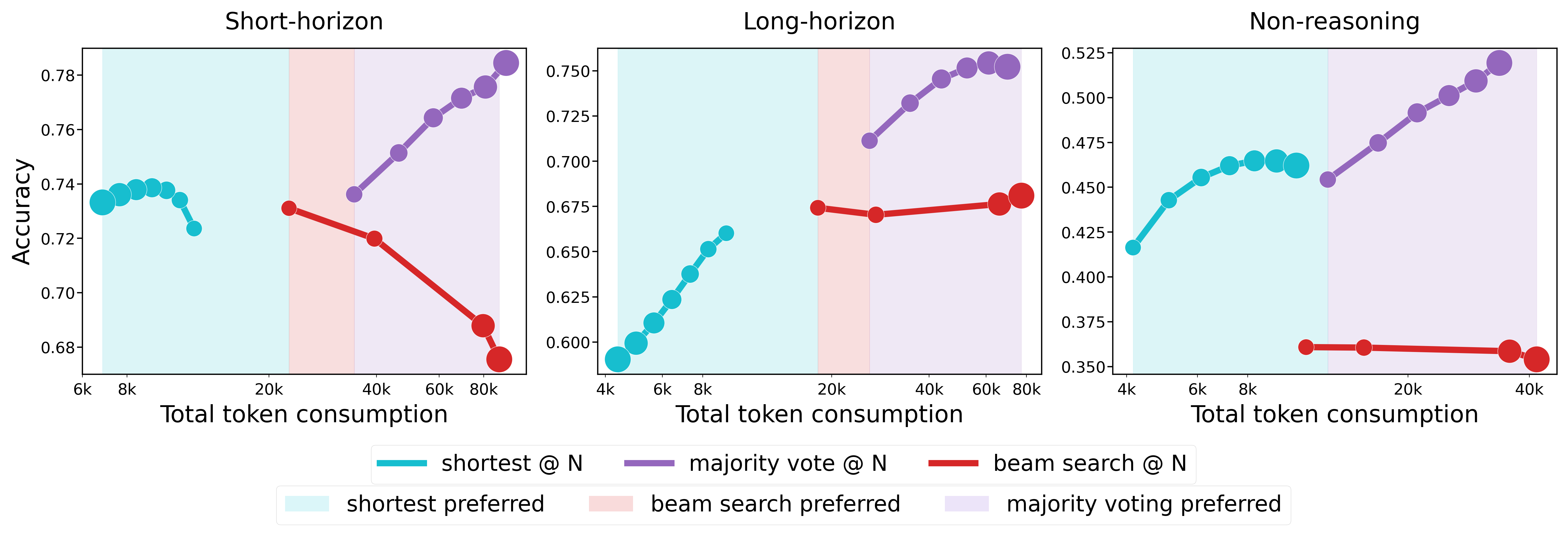}
    \caption{Plots of shortest (cyan), majority-voted (purple), and beam-searched (red) trace performances for short-horizon (left), long-horizon (middle), and non-reasoning (right) models. Short-horizon models include \textit{R1}, \textit{DAPO-32B}, and \textit{QwQ-32B}; long-horizon models include  and \textit{Qwen3-32B}, \textit{GPT-OSS-120B} and \textit{R1-32B}; and non-reasoning models include \textit{Qwen3-235B-Instruct} and \textit{DeepSeek-Chat}. Performance is measured using average accuracy on the AIME 2024--2025 and GPQA Diamond datasets. Shaded regions show the optimal TTS strategy by compute budget: shortest for low compute, beam search for medium, majority voting for high. The plot illustrates that there is no free lunch for TTS strategies: no single strategy is optimal and optimality depends on compute budget. This highlights the need for a principled, model-aware approach to determine the best scaling strategy at test-time. Marker size increases with N ($N \geq 2$); N is the number of parallel traces sampled.}
    \label{fig:teaser}
\end{figure}

\section{Introduction}

Test-time scaling (TTS) has emerged as an effective method to enhance the reasoning capabilities of large language models (LLMs) by increasing inference-time compute. Any sound TTS strategy, by definition, will exhibit performance improvements as more compute is allocated. The best TTS strategy to choose, however, remains an open question.

Early studies explored sequential scaling methods, either by artificially extending reasoning traces~\citep{muennighoff2025s1simpletesttimescaling} or by encouraging deeper exploration within a single reasoning direction before switching, thereby mitigating ``underthinking''~\citep{wang2025thoughtsplaceunderthinkingo1like}. More recent analyses have questioned the benefits of such sequential extensions. Notably, \citet{gema2025inversescalingtesttimecompute} examined synthetic tasks designed to isolate specific reasoning abilities, such as counting with distractors, and regression with spurious correlations. Their findings indicate that longer reasoning can reinforce incorrect behaviors, amplify errors, and misalign reasoning paths, thereby degrading accuracy and even introducing safety concerns.

Similarly, \citet{hassid2025dontoverthinkitpreferring} proposed \textit{short-m@k}, a parallel TTS technique where the final prediction is obtained by majority voting among the $m$ shortest reasoning traces, out of $k$ sampled outputs. Their results support the idea that shorter, more concise reasoning often outperforms extended deliberation.

Prior studies, while offering valuable insights, do not account for model variations and rely on older reasoning models. In this work, we revisit these findings using more recent models, including the GPT-OSS~\citep{openai2025gptoss120bgptoss20bmodel} and Qwen3~\citep{yang2025qwen3technicalreport} series. Our findings reveal that the relationship between compute and performance varies across model families—a divergence we attribute to differences in their post-training algorithms.

We argue that distinct post-training methods give rise to varying \textbf{reasoning horizons}. Models with large horizons (large-horizon models) are able to sustain deeper reasoning by means of longer traces, thereby benefiting in performance on harder tasks where greater thought is necessary. Short-horizon models, however, cannot generate long coherent traces, thereby making it most suitable for them to prioritize concise reasoning, irrespective of problem difficulty.

As a consequence of their training dynamics, short-horizon models commonly emerge from post-training with GRPO or GRPO-like algorithms, aligning with the well-documented length bias introduced by GRPO~\citep{yu2025dapoopensourcellmreinforcement}. In contrast, long-horizon models are typically produced by alternative reinforcement-learning methods that maintain stability over extended traces. For example, Qwen3—a long-horizon model—is post-trained using GSPO rather than GRPO~\citep{zheng2025groupsequencepolicyoptimization}. This observation supports our hypothesis that the choice of post-training strategy plays a key role in determining a reasoning model’s effective horizon.

Overall, our work highlights the need for a model-aware perspective on TTS that accounts for differences in training methodology, problem difficulty, and compute availability to guide principled strategy selection.

%% file: sections/preliminaries.tex
\section{Preliminaries}
\subsection{Test-time scaling methods}

\begin{wrapfigure}{r}{0.5\textwidth}
    \vspace{-10pt}
    \centering
    \includegraphics[width=0.48\textwidth]{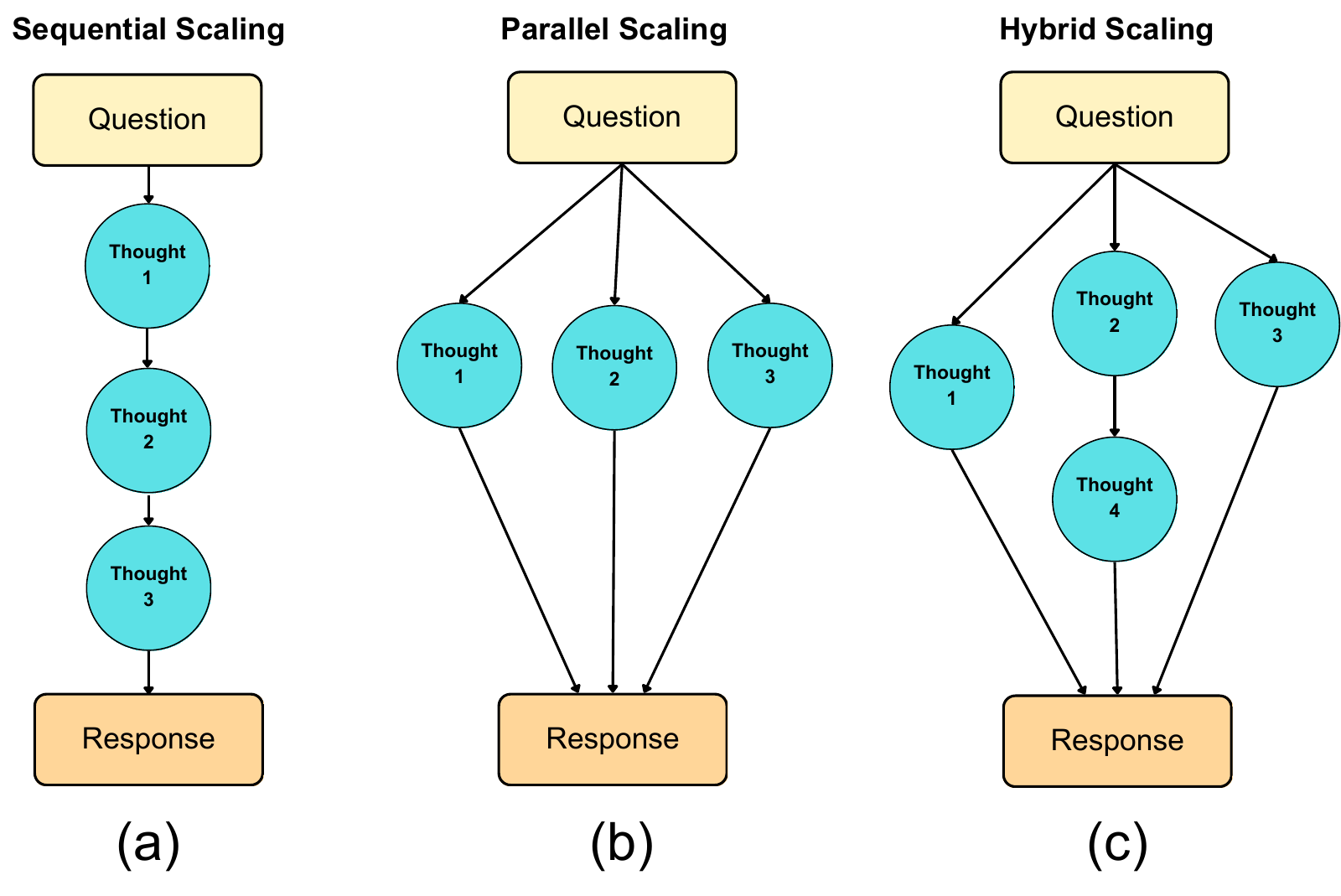}
    \vspace{-5pt}
    \caption{Different TTS paradigms}
    \vspace{-5pt}
    \label{fig:tts_techniques}
\end{wrapfigure}

Test-time scaling strategies for LLMs vary widely, typically falling into parallel, sequential, hybrid/meta, and internal compute mechanisms (Figure~\ref{fig:tts_techniques}). While each class of methods shows promise in specific settings, no single strategy is universally optimal. 

\textbf{Parallel scaling strategies} improve performance by aggregating answers across multiple independently sampled reasoning traces. Self-consistency~\citep{wang2023selfconsistencyimproveschainthought} samples diverse reasoning paths and chooses the most frequent final answer, significantly improving performance on arithmetic and symbolic tasks. Best-of-$n$ sampling is widely used as a simple parallel method~\citep{snell2024compute}, though more principled voting strategies like majority voting~\citep{lightman2023verify}, and Multi-Agent Verification (MAV)~\citep{lifshitz2025multi} have been recently proposed. Short-m@k~\citep{hassid2025dontoverthinkitpreferring} exploits early stopping: it runs $k$ reasoning chains in parallel and halts early based on the proportion of traces completed.

\textbf{Sequential scaling strategies} extend reasoning depth by iteratively refining, restarting, or backtracking. Chain-of-Thought (CoT) prompting~\citep{wei2023chainofthoughtpromptingelicitsreasoning} is a fundamental idea, and subsequent work like STaR~\citep{zelikman2022star} and Reflexion~\citep{shinn2023reflexion} explore revision through trial-and-error or verbal self-reflection. Tree-of-Thought (ToT)~\citep{yao2023tree} and Graph-of-Thoughts~\citep{besta2024graph} scale this further via structured breadth-first or DAG-style search. \mbox{AlphaGeometry~\citep{chervonyi2025goldmedalistperformancesolvingolympiad}} integrates symbolic proof search with LLMs for step-level sequential control. S1~\citep{muennighoff2025s1simpletesttimescaling} fine-tunes models for teaching self-correction strategies, utilizing higher test-time compute. 

More recent efforts like \textbf{hybrid scaling strategies} blend both axes. Meta-Reasoner~\citep{sui2025metareasoner} uses contextual bandits to dynamically select TTS strategies based on perceived task difficulty. AgentTTS~\citep{wang2025agenttts} and START~\citep{li2025start} deploy agents (LLMs with tool-calling capabilities) to switch between direct generation or more elaborate reasoning. PEARL~\citep{liu2025pearlparallelspeculativedecoding} interleaves draft generation with refinement, simulating self-improvement loops. These meta-schedulers recognize that neither deep nor parallel scaling alone is enough, and aim to adapt the strategy based on model behavior and prompt dynamics. Internal scaling strategies, in contrast, modify how much computation the model performs internally during inference, without explicitly adjusting the number of external samples or reasoning steps. HALT-CoT~\citep{laaouach2025haltcot} and SoftCoT++~\citep{xu2025softcottesttimescalingsoft} estimate answer uncertainty and terminate early if confidence is high. 

\paragraph{No strategy is universally best.} Multiple empirical studies reinforce that no TTS strategy consistently dominates. ~\citet{zhang2025survey} emphasize tradeoffs across accuracy, consistency, and efficiency—the ``TTS trilemma.'' ~\citet{snell2024compute} show that compute-optimal allocation (e.g., short inference on easy questions, deeper inference on hard ones) outperforms scaling model size alone. ~\citet{ghosal2025doesthinkinghelpmirage} and ~\citet{hassid2025dontoverthinkitpreferring} show that longer CoT chains often degrade accuracy. Inverse-scaling effects~\citep{gema2025inversescalingtesttimecompute} demonstrate that larger models or longer prompts may hurt, especially when uncertainty is high or symbolic reasoning is required. This underscores our central thesis: optimal TTS is highly contextual and must consider model training (e.g., type of post-training), task type, and difficulty.

In this work, we consider first finish search (FFS, Algorithm~\ref{alg:ffs}), last finish search (LFS, Algorithm~\ref{alg:lfs}) and beam search for our analyses, the first two of which are parametrized by variables k and N, while the last is parametrized by N alone. FFS-k@N means sampling N outputs and performing MV among the shortest k samples to determine the majority vote while LFS-k@N simply involves choosing the longest k samples instead of shortest, followed by majority voting on these. Beam search involves maintaining a beam of high probability partial hypotheses, continuously updating these prefixes as decoding progresses.\footnote{Since we use API-based model inference (\verb|deepinfra.com|), we restrict our analysis to API-friendly TTS strategies.}

\begin{figure*}[t]
\centering



\begin{minipage}[t]{0.48\textwidth}
\begin{algorithm}[H]
\footnotesize
\caption{First Finish Search - k (FFS-k)}
\begin{algorithmic}[1]
\Require Model $M$, prompt $x$, number of samples $N$, filter size $k$
\Ensure Final answer $y^*$
\State Generate $N$ outputs $\{y_1, \dots, y_N\}$ in parallel
\State Stop as soon as $k$ traces are complete
\State Select these $k$ traces $\{y_{(1)}, \dots, y_{(k)}\}$
\State Extract final answers from these $k$ traces
\State \Return majority-voted answer among them
\vspace{10pt}
\end{algorithmic}
\label{alg:ffs}
\end{algorithm}
\end{minipage}
\hfill
\begin{minipage}[t]{0.48\textwidth}
\begin{algorithm}[H]
\footnotesize
\caption{Last Finish Search - k (LFS-k)}
\begin{algorithmic}[1]
\Require Model $M$, prompt $x$, number of samples $N$, filter size $k$
\Ensure Final answer $y^*$
\State Generate $N$ outputs $\{y_1, \dots, y_N\}$ in parallel
\State Sort \textbf{completed} outputs by trace length (descending)
\State Select longest $k$ traces $\{y_{(1)}, \dots, y_{(k)}\}$
\State Extract final answers from these $k$ traces
\State \Return majority-voted answer among them
\end{algorithmic}
\label{alg:lfs}
\end{algorithm}
\end{minipage}
\vspace{-10pt}
\end{figure*}

\subsection{Models}

We evaluate both reasoning and non-reasoning models to analyze the effects of TTS strategies across diverse training paradigms.

\paragraph{Reasoning models.} \textbf{DeepSeek-R1} is a reasoning-tuned LLM optimized for mathematical and logical tasks using GRPO—an RL algorithm that improves efficiency over PPO but introduces biases in gradient normalization, leading to uneven penalization across trace lengths. \textbf{R1-32B} is a distilled 32B-parameter variant of DeepSeek-R1 that inherits its reasoning-centric behavior, exhibiting similar trace-length-dependent trends at reduced capacity. \textbf{QwQ-32B} is a reasoning-focused model from Qwen that leverages stronger MoE routing, typically producing shorter and more compact reasoning traces. \textbf{GPT-OSS-120B} is a large open-source GPT-style model trained with extensive reasoning supervision, serving as a transparent large-scale baseline. \textbf{Qwen3-32B} belongs to the Qwen3 family and emphasizes diverse reasoning domains—STEM, code, and commonsense—yielding qualitatively distinct reasoning patterns from DeepSeek models. \textbf{DAPO-32B} is a RL-trained reasoning model based on the DAPO algorithm, an open-source alternative to GRPO that claims to mitigate its gradient normalization bias while maintaining sample efficiency.

\paragraph{Non-reasoning models.} \textbf{Qwen3-235B-Instruct} is a large instruction-tuned MoE model (235B total, 22B active parameters) without explicit reasoning supervision, producing fluent but unstructured responses. \textbf{DeepSeek-Chat} is the general-purpose conversational model from DeepSeek, optimized for dialogue and summarization rather than multi-step reasoning, allowing us to assess the impact of TTS on models without reasoning-centric training. 

\subsection{Datasets}

We evaluate models on two complementary reasoning benchmarks—\textbf{AIME} and \textbf{GPQA Diamond}—which together cover both symbolic-numerical and conceptual reasoning domains.

The \textbf{American Invitational Mathematics Examination (AIME)} is a high-school level contest assessing symbolic and arithmetic reasoning through 30 short-answer problems, each with an integer solution between 0 and 999. We use three recent variants—\textbf{AIME 2024}, \textbf{AIME 2025-I}, and \textbf{AIME 2025-II}—to test consistency across different years and question distributions. Each problem is formatted as a concise natural-language prompt, and models are instructed to output the final answer within ``\texttt{\textbackslash boxed{}}'' for consistent evaluation. AIME problems typically require multi-step deductive reasoning, involving algebraic or combinatorial manipulation, making them ideal for analyzing the accuracy-efficiency trade-offs of TTS strategies.

\textbf{GPQA Diamond}~\citep{rein2023gpqagraduatelevelgoogleproofqa} is a graduate-level benchmark designed to test conceptual and factual reasoning across physics, biology, and chemistry. Each question is multiple-choice with four options (A--D), and models must output the selected answer in a ``\texttt{\textbackslash boxed{}}'' format for standardized parsing.
We employ the \textbf{Diamond} subset, the most challenging and expert-verified split, emphasizing high conceptual depth and factual precision.
In contrast to AIME's numerical reasoning focus, GPQA Diamond evaluates abstract and knowledge-grounded reasoning. Together, they provide a comprehensive view of reasoning performance across mathematical, symbolic, and conceptual domains.

All model- and dataset-specific hyperparameters are listed in Appendix~\ref{appx:hyper}.

\subsection{Metrics}

\paragraph{Accuracy.}


This metric is defined as the proportion of generated traces whose final prediction matches the ground-truth answer. Even when parsing is reliable, and there is only a single gold answer, representational ambiguities can arise. For instance, the fraction $\frac{1}{2}$ may be written as ``1/2'' or ``0.5,'' both of which are semantically equivalent but syntactically distinct. In our setting, accuracy evaluation is simplified because three of the four datasets are derived from AIME, where answers are restricted to three-digit integers. The remaining dataset (GPQA) is multiple-choice with options limited to A--D. For both cases, we explicitly instruct the model to provide its final answer within delimiters, which facilitates reliable extraction of the predicted value.

\paragraph{Token consumption.}

We count token consumption in two distinct ways, capturing different aspects of the utilized compute. \textbf{Total tokens} refer to the total number of tokens generated across all the traces in order to arrive at the generated answer. \textbf{Sequential tokens}, on the other hand, refer to the number of tokens that must necessarily be produced in a sequence, and are dependent on the previously generated tokens. For instance, during vanilla decoding using greedy or stochastic sampling, generating a trace $x$ would involve a sequential token count of $|x|$ since $x_i$ can only be generated after all of $X_j$, $j < i$ are generated. For an inference strategy which requires generating N complete traces $x^1$, $x^2$, ..., $x^N$, the sequential token count would be $\max^N_{i=1}{|x^i|}$ since all the tokens in the longest trace would be dependent on the one before it. Therefore, while total tokens measure the overall compute used, sequential token count gives an estimate of the minimum possible latency in generating a given output (assuming each token generation takes the same time).

%% file: sections/prob_diff.tex
\subsection{Measuring Problem Difficulty}
\label{sec:prob-diff}

\begin{figure}[t]
    \centering
    \begin{subfigure}[t]{0.24\textwidth}
        \centering
        \includegraphics[width=\linewidth]{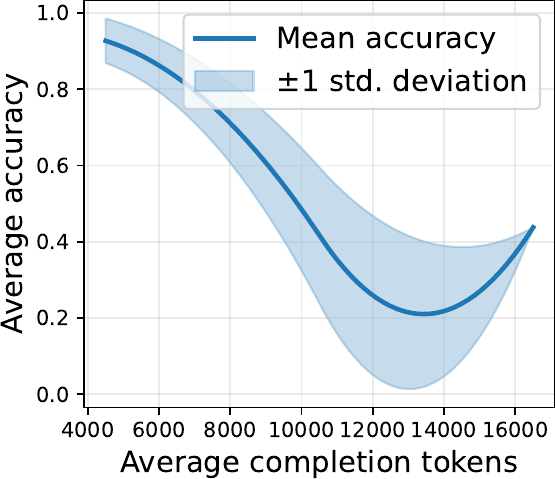}
            \caption{AIME2024}
    \end{subfigure}
    \hfill
    \begin{subfigure}[t]{0.24\textwidth}
        \centering
        \includegraphics[width=\linewidth]{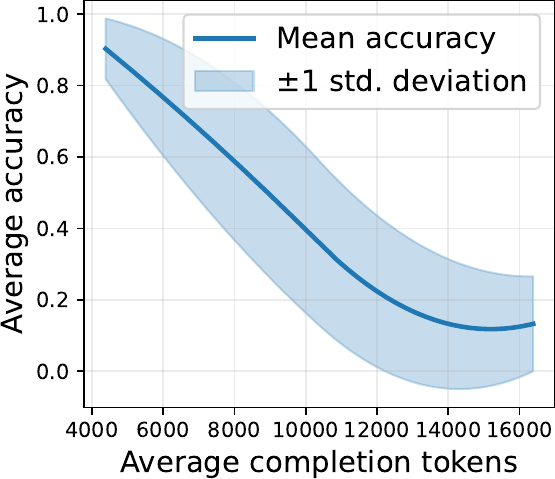}
        \caption{AIME2025-I}
    \end{subfigure}
    \hfill
    \begin{subfigure}[t]{0.24\textwidth}
        \centering
        \includegraphics[width=\linewidth]{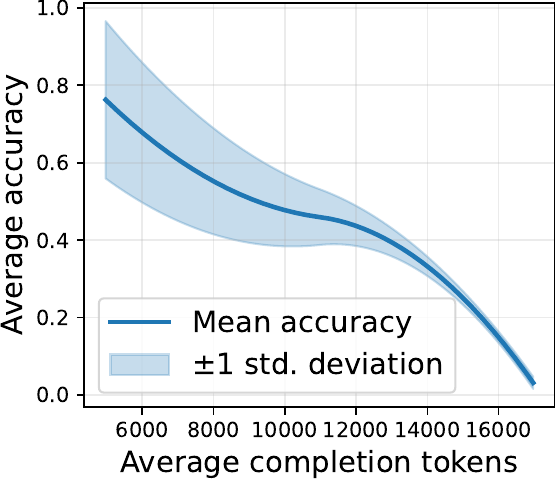}
        \caption{AIME2025-II}
    \end{subfigure}
    \hfill
    \begin{subfigure}[t]{0.24\textwidth}
        \centering
        \includegraphics[width=\linewidth]{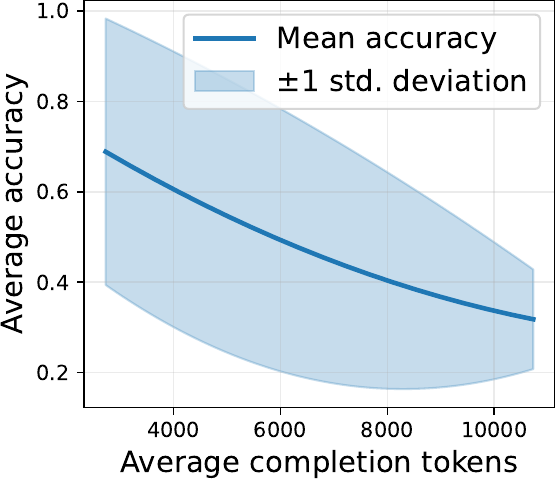}
        \caption{GPQA Diamond}
    \end{subfigure}
    \caption{Mean accuracy vs.\ average completion tokens for different datasets averaged across all models}
    \label{fig:acc_vs_tokens}
\end{figure}

In order to devise a granular recipe for the appropriate scaling strategy to use at test-time, it is crucial to take into account the problem difficulty. The direct way to measure difficulty of a question is to simply calculate the task accuracy for a given problem, averaged across all models and sampled traces. Another, more indirect approach would be to calculate the average tokens generated for the task, again averaged across all models and outputs. Interestingly, we find that both these metrics are correlated (Figure~\ref{fig:acc_vs_tokens}), and that this overall trend holds across all datasets, where a reasoning (as well as a non-reasoning) models ``thinks'' longer on harder (as measured through accuracy) problems. 

\begin{findings}
Both reasoning and non-reasoning models think longer for harder problems.
\end{findings}

While both reasoning and non-reasoning models expend more tokens on harder problems, longer trace lengths alone do not guarantee improved quality. Recent research suggests that excessive deliberation can harm performance by propagating early mistakes, contributing to the growing view that more compute is not always better~\citep{gema2025inversescalingtesttimecompute}.

%% file: sections/results.tex
\section{Results}

\subsection{Beam search shows inverse or no scaling}


We notice that across two of the model families—short-horizon and non-reasoning—beam search exhibits a consistent inverse-scaling pattern: \emph{performance degrades monotonically as the beam size N increases} (Figure~\ref{fig:teaser}). For short-horizon models such as R1 and QwQ-32B, accuracy drops sharply once N becomes larger than 2; for non-reasoning models there is a similar, although milder, trend of performance drops as N increases. Even long-horizon models like GPT-OSS-120B, and Qwen3-32B fail to benefit from beam expansion: their accuracy curves flatten or decline as N increases. Since total token consumption—and therefore total compute—increases with beam width, these results reveal a clear case of \emph{inverse compute scaling}, where allocating more test-time compute via larger beams either harms accuracy or yields no benefit.

\begin{findings}
    Beam search performance degrades or remains the same with increasing beam size for reasoning-focused datasets like AIME and GPQA.
\end{findings}

\subsection{Correlation of trace length with quality}
\label{sec:trace-len-quality}

It is crucial to understand how the trace length correlates with quality (as measured through accuracy) in order to obtain a deeper understanding of length-based filtering strategies like FFS and LFS. FFS and LFS are based on two diametrically opposite viewpoints: \textit{shorter is better} and \textit{longer is better}. To investigate which hypothesis (or hypotheses) hold for a given model, we report the accuracy for a given interval of trace lengths and problem difficulties (Table~\ref{tab:combined_models}). Note that the problem difficulty is measured by averaging the accuracy over all models and traces (Section~\ref{sec:prob-diff}), while the reported accuracy is measured by averaging over all outputs for the specific model. A key consideration is that problem difficulty is confounded with trace length (Figure~\ref{fig:acc_vs_tokens}): short traces typically arise from easier problems, whereas long traces tend to correspond to harder ones. To mitigate this confounding effect, we restrict our analysis to tasks for which both short and long traces are available. For each such dataset, we compute a single accuracy value for short and long traces separately, and then average these values across datasets, thereby preventing differences in dataset size from disproportionately influencing the aggregated results. Based on the ordering between these reported accuracies, we broadly classify the six reasoning models as either short-horizon or long-horizon. While the two non-reasoning models both show short-horizon behavior, we choose to keep them separate from short-horizon models due to the significant differences in the post-training techniques employed (instruction tuning vs.\ RL).

\begin{table}[t!]
\centering
\caption{Model categorization, behavioral characteristics, and accuracy as a function of trace length and problem difficulty. Tasks are classified as easy or hard based on whether their difficulty is below or above the median across all tasks. Trace lengths are labeled short or long using the model-specific median trace length computed over the entire task set.}
\label{tab:combined_models}
\resizebox{\textwidth}{!}{%
\begin{tabular}{llp{0.32\textwidth}cccc}
\toprule
\multirow{4}{*}{\textbf{Category}} &
\multirow{4}{*}{\textbf{Model}} &
\multirow{4}{*}{\textbf{Behavior}} &
\multicolumn{4}{c}{\textbf{Accuracy}} \\
\cmidrule(lr){4-7}
& & & \multicolumn{2}{c}{\textbf{Easy}} &
\multicolumn{2}{c}{\textbf{Hard}} \\
\cmidrule(lr){4-5} \cmidrule(lr){6-7}
& & & \textbf{Short} & \textbf{Long} & \textbf{Short} & \textbf{Long} \\
\midrule

\multirow{3}{*}{Short horizon}
 & R1 & \multirow{3}{=}{Shorter is always better}
   & 0.95 & 0.72 & 0.61 & 0.48 \\

 & DAPO-32B &  & 0.80 & 0.54 & 0.05 & 0.05 \\

 & QwQ-32B  &  & 0.91 & 0.70 & 0.58 & 0.58 \\
\midrule

\multirow{3}{*}{Long horizon}
 & GPT-OSS-120B & \multirow{3}{=}{Shorter is better for easy problems while longer is better for hard problems}
   & 0.92 & 0.85 & 0.48 & 0.53 \\

 & Qwen3-32B &  & 0.75 & 0.63 & 0.22 & 0.45 \\

 & R1-32B   &  & 0.92 & 0.62 & 0.33 & 0.34 \\
\midrule

\multirow{2}{*}{Non-reasoning}
 & Qwen3-235B & \multirow{2}{=}{Shorter is always better}
   & 0.90 & 0.52 & 0.51 & 0.20 \\

 & DeepSeek   &  & 0.47 & 0.22 & 0.12 & 0.06 \\
\bottomrule
\end{tabular}%
}
\end{table}

Across all models, we observe a consistent invariant: for any given trace-length bucket, the reported accuracy is always higher on easy problems than on hard ones. This pattern is expected, as problem difficulty is defined through aggregated accuracy, and harder questions naturally exhibit lower correctness rates.

It is more interesting to observe how the order between short and long traces for easy and hard problems varies across different models. For short-horizon models (R1, QwQ-32B, DAPO-32B), we find that for a given problem difficulty, shorter traces are more likely to be correct than longer ones. This is in line with the recent observations made by \citep{agarwal2025finishsearchefficienttesttime,hassid2025dontoverthinkitpreferring} where the authors find that conciseness in the reasoning trace is linked to better accuracy. However, we observe a different phenomenon with other more advanced models such as Qwen3-32B and GPT-OSS-120B, where for easier problems shorter traces are better, but for harder problems longer traces are preferred.

DAPO-32B shows a similar length bias pattern to prior models, with shorter traces more likely to be correct than longer ones (Table~\ref{tab:combined_models}). The bias level is also close to that of R1, which suggests that any improvements in mitigating length bias over GRPO may be limited under our evaluation.

The complete results for different models, datasets, and TTS strategies can be found in Appendix~\ref{appx:result}. Individual, model-wise plots for FFS and LFS are present in Appendix~\ref{appx:plots}.

\begin{findings}
    DAPO induces length bias to the same extent as GRPO.
\end{findings}

%% file: sections/analysis.tex
\begin{figure}[t]
    \centering
    \includegraphics[width=\linewidth]{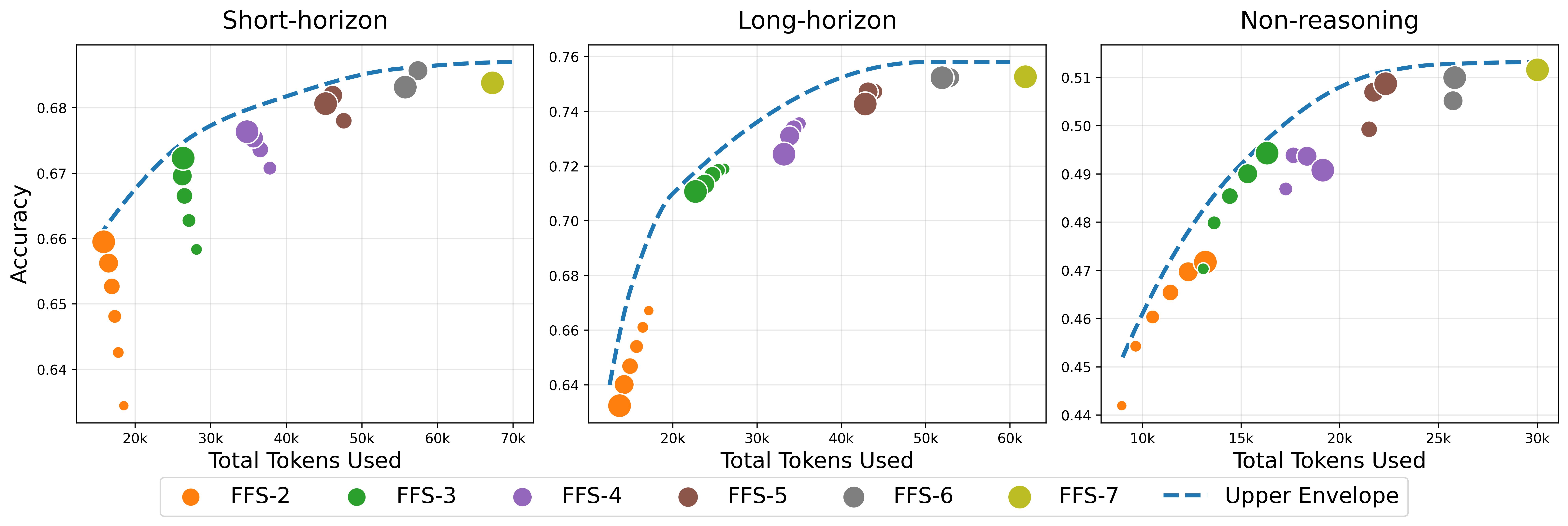}
    \caption{Accuracy versus token usage for different model families. FFS-k variants are shown in distinct colors (one color per k). Marker size encodes the value of N, with larger markers representing larger N.}
    \label{fig:ffs-closer-look}
\end{figure}

\vspace{-6pt}

\begin{figure}[t]
    \centering
    \includegraphics[width=\linewidth]{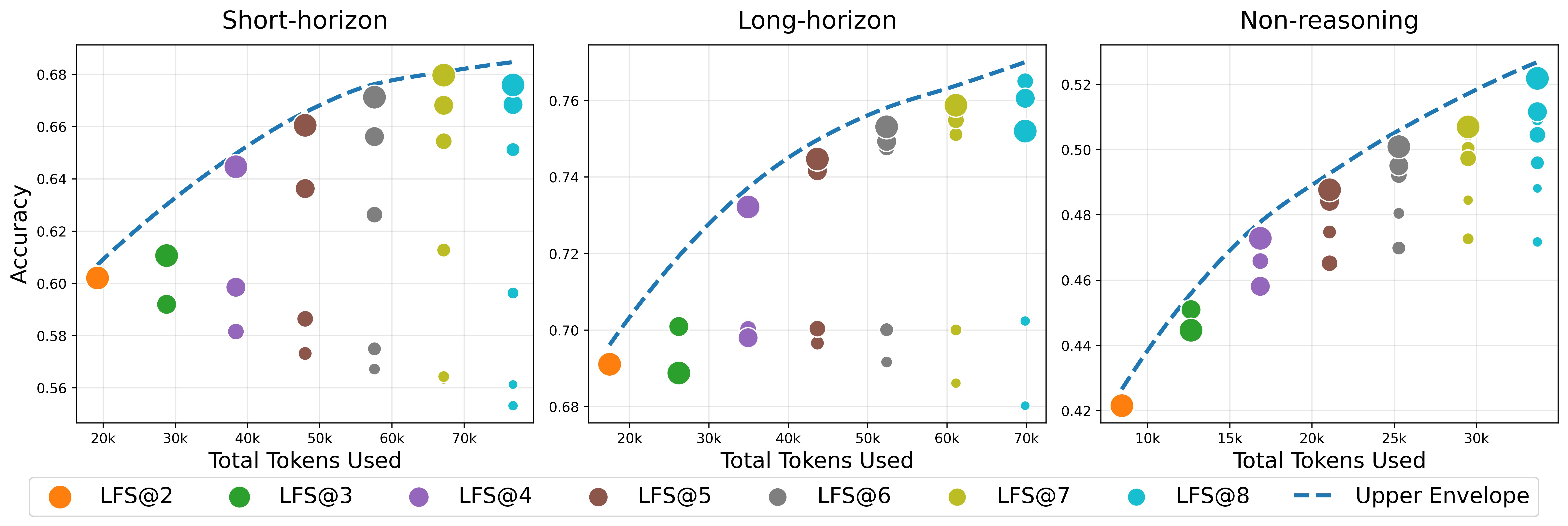}
    \caption{Accuracy versus token usage for different model families. LFS@N variants are shown in distinct colors (one color per N). Marker size encodes the value of k, with larger markers representing larger k.}
    \label{fig:lfs-closer-look}
\end{figure}

\section{Analysis}
\label{sec:analysis}

It is necessary to determine how the performance of FFS-k@N and LFS-k@N varies for different values of k and N across the models, in order to find the optimal strategy. Figures~\ref{fig:ffs-closer-look} and \ref{fig:lfs-closer-look} depict the performance of FFS-k@N and LFS-k@N for the different model types. These plots reveal an interesting behavior where the optimal TTS strategy always seems to scale with increasing budget. 

Furthermore, we find that for the LFS family of methods, the maximum performance for a given amount of total compute is always achieved when k is large (which implies k=N). Note that k=N is simply MV-N, and therefore we conclude that MV@N is better than LFS-k@N for any value of k, all while consuming the same number of tokens.

\begin{findings}
LFS is always suboptimal to MV: longest‐trace filtering consistently reduces accuracy at the same compute.
\end{findings}

For the FFS family of methods, we observe a more nuanced behavior where while performance improves for increasing k (while also consuming more tokens) across all model types, the behavior with N for a fixed k is mixed. We find that for short-horizon models larger values of N are always best (higher performance at lesser token consumption), while for long-horizon and non-reasoning models there is a tradeoff between performance and the compute consumed. Note that while a tradeoff exists for both long-horizon and non-reasoning models, the handles to vary the tradeoff are opposite for them: for long-horizon models to draw performance at the cost of higher compute one has to choose smaller N (essentially performing simple decoding) while for non-reasoning models one has to choose larger N.

%% file: sections/recipe.tex
\section{The Recipe}
\definecolor{rowgray}{gray}{0.96}

\begin{table}[t]
\centering
\captionsetup{
    labelfont=bf,
    font=small,
    skip=6pt
}
\setlength{\tabcolsep}{8pt}
\begin{tabular}{
    >{\raggedright\arraybackslash}m{2.6cm}
    >{\centering\arraybackslash}m{2.5cm}
    >{\centering\arraybackslash}m{2.5cm}
    >{\raggedright\arraybackslash}m{4cm}
}
\toprule
\rowcolor{rowgray}
\textbf{Model Family} & \textbf{Difficulty} & \textbf{Compute} & \textbf{Recommended Recipe} \\
\midrule

\multirow{2}{*}{\textbf{Short-horizon}} 
 & High / Low & High & MV@N; N large \\
 & High / Low & Low & FFS-k@N; k=1, N large \\

\midrule
\multirow{2}{*}{\textbf{Long-horizon}} 
 & High / Low & High & MV@N; N large \\
 & High / Low & Low & SD \\
 
\midrule
\multirow{2}{*}{\textbf{Non-reasoning}} 
 & High / Low & High & MV@N; N large \\
 & High / Low & Low & FFS-k@N; k=1, N large \\

\bottomrule
\end{tabular}
\caption{
Decision matrix outlining optimal TTS strategies based on model family, task difficulty, and computational budget. K denotes the number of shortest/longest traces considered for voting, and N indicates the total trace count. SD refers to simple decoding, a greedy left-to-right generation procedure analogous to beam search with beam size 1: at each generation step, the model selects only the single most probable continuation.
\vspace{-7pt}
}
\label{tab:training_recipe_final}
\end{table}

Our analysis reveals that the optimal test-time scaling strategy is not universal but depends on a combination of the model's architectural family, the difficulty of the problem at hand, and the available compute budget. To distill our findings into actionable guidance, we present a decision matrix in Table~\ref{tab:training_recipe_final}. We explain below the rationale for choosing such a recipe below.

\paragraph{Short-horizon models.}
Across both low- and high-difficulty settings, short-horizon models consistently prefer shorter traces over longer ones (Table~\ref{tab:combined_models}). Because FFS-k improves in both accuracy and computational cost as $k$ increases, we select small values of $k$ (specifically $k = 1$) under low-compute constraints, and large values of k (namely $k = N$) when ample compute is available. The latter choice is equivalent to MV@N, since selecting the $k = N$ shortest traces from N samples necessarily involves including all traces. Additionally, for short-horizon models, performance increases with larger N for any fixed k. Accordingly, we choose N to be as large as permitted by the compute budget.

\paragraph{Long-horizon models.}
For high-difficulty settings, long-horizon models prefer longer traces. Because LFS@N improves as N increases, we use large N when compute is abundant and small N (ideally $N = 1$) when compute is limited. Keeping N fixed, performance increases with larger k; thus, in both compute regimes we set $k = N$, which corresponds to MV@N. Under low-compute conditions where $N = 1$, MV@N reduces to simple decoding (SD) without any aggregation. 

For low-difficulty settings, we instead use FFS, since long-horizon models prefer shorter traces for easier problems. As with short-horizon models, FFS-k scales positively with $k$, so we employ a large $k$ when compute is high and a small k when compute is low. In these settings, performance improves as N decreases (in contrast to short-horizon models, where larger N is beneficial). Therefore, we set $N = k$, which yields the MV@N strategy. Under low compute, this results in $k = 1$ and thus simple decoding (SD), while under high compute it corresponds to MV with a large sample size.

Interestingly, although the model types exhibit distinct behavior across different task difficulties, the optimal TTS strategy is ultimately independent of the problem difficulty, as shown in the final recipe (Table~\ref{tab:training_recipe_final}). 

\begin{findings}
The optimal TTS strategy is independent of task difficulty.
\end{findings}

%% file: sections/conclusion.tex
\section{Conclusion} Our large-scale study demonstrates there is no single optimal test-time scaling (TTS) strategy for enhancing LLM reasoning. The most effective approach is contingent on a crucial interplay between the model's training methodology, problem difficulty, and the available compute budget. We find that different model families exhibit distinct behaviors: \textbf{short-horizon} models consistently favor shorter, concise traces, while \textbf{long-horizon} models benefit from longer, more deliberate reasoning for harder problems while concise reasoning for easier problems. Critically, beam search consistently proves suboptimal for complex reasoning. Our work provides a practical framework for practitioners, underscoring that maximizing performance requires a nuanced, model-aware approach to inference rather than a universal strategy.

%% file: sections/appendix.tex
\appendix

\section{Hyperparameters}
\label{appx:hyper}

All hyperparameters for our experiments are given in Table~\ref{tab:hyperparams_all}.

\begin{table}[h]
  \centering
    \resizebox{\linewidth}{!}{%
  \begin{tabular}{lcccccc}
    \toprule
    \textbf{Model} & \textbf{GPQA} & \textbf{AIME24} & \textbf{AIME25-I} & \textbf{AIME25-II} & \textbf{Top-$p$} & \textbf{Temp.} \\
    \midrule
    Deepseek              & 16K & 32K & 32K & 32K & 0.95 & 0.6 \\
    R1                & 32K & 32K & 32K & 32K & 0.95 & 0.6 \\
    QwQ               & 32K & 32K & 32K & 32K & 0.95 & 0.6 \\
    R1-Distill-Qwen   & 32K & 32K & 32K & 32K & 0.95 & 0.6 \\
    GPT-OSS-120B      & 8K  & 8K  & 8K  & 8K  & 0.95 & 0.6 \\
    Qwen3-235B        & 5K  & 5K  & 5K  & 5K  & 0.95 & 0.6 \\
    Qwen3             & 16K & 32K & 32K & 32K & 0.95 & 0.6 \\
    Dapo-Qwen-32B     & 10.1K & 20.5K & 20.5K & 20.5K & 0.7 & 1.0 \\
    \midrule
    \multicolumn{7}{c}{\emph{Global settings (shared across all models)}} \\[2pt]
    Beam width                 & \multicolumn{6}{c}{8} \\
    Samples ($n$, MV/LFS/FFS)  & \multicolumn{6}{c}{8} \\
    Answer-reserve for BF      & \multicolumn{6}{c}{3K} \\
    \bottomrule
  \end{tabular}
  }
  \caption{Decoding hyperparameters used in all experiments across models and datasets. 
           Identical values across datasets are shown once.}
  \label{tab:hyperparams_all}
\end{table}

\section{Results}
\label{appx:result}

\begin{table*}[h]
\centering
\label{tab:all_results}

\setlength{\tabcolsep}{2pt}
\renewcommand{\arraystretch}{1.08}

\begin{tabular}{@{}c@{}}
\begin{minipage}[t]{0.31\linewidth}
\centering\small
\resizebox{!}{0.31\linewidth}{
\begin{tabular}{lcccc}
\toprule
Metric & BS & MV & LFS & FFS \\
\midrule
Seq. tokens & -- & 9.2k & 9.2k & \best{3.0k} \\
Total tokens & -- & 45.4k & 45.4k & \best{4.3k} \\
\hdashline
GPQA & -- & 54.0 & 53.5 & \best{55.1} \\
AIME24 & -- & \best{53.3} & 46.7 & \best{53.3} \\
AIME25-I & -- & \best{46.7} & 40.0 & 33.3 \\
AIME25-II & -- & \best{46.7} & 33.3 & 40.0 \\
\bottomrule
\end{tabular}
}
\vspace{-5pt}
\caption*{(a) Dapo-Qwen-32B}
\end{minipage}
\vspace{4pt}
\quad
\begin{minipage}[t]{0.31\linewidth}
\centering\small
\resizebox{!}{0.31\linewidth}{
\begin{tabular}{lcccc}
\toprule
Metric & BS & MV & LFS & FFS \\
\midrule
Seq. tokens & 6.3k & 13.3k & 13.3k & \best{0.8k} \\
Total tokens & 50.3k & 31.6k & 31.6k & \best{0.9k} \\
\hdashline
GPQA & 56.6 & \best{58.1} & 53.0 & 48.5 \\
AIME24 & 26.7 & \best{43.3} & 36.7 & 26.7 \\
AIME25-I & 26.7 & \best{46.7} & 33.3 & 33.3 \\
AIME25-II & 20.0 & 20.0 & \best{26.7} & 13.3 \\
\bottomrule
\end{tabular}
}
\vspace{-5pt}
\caption*{(b) Deepseek-Chat}
\end{minipage}
\vspace{4pt}
\quad
\begin{minipage}[t]{0.31\linewidth}
\centering\small
\resizebox{!}{0.31\linewidth}{
\begin{tabular}{lcccc}
\toprule
Metric & BS & MV & LFS & FFS \\
\midrule
Seq. tokens & 4.4k & 5.0k & 5.0k & \best{3.0k} \\
Total tokens & 34.8k & 32.0k & 32.0k & \best{5.9k} \\
\hdashline
GPQA & 71.7 & \best{72.2} & 66.2 & 66.2 \\
AIME24 & 73.3 & \best{80.0} & \best{80.0} & 76.7 \\
AIME25-I & 60.0 & \best{80.0} & 73.3 & 66.7 \\
AIME25-II & 80.0 & \best{86.7} & \best{86.7} & \best{86.7} \\
\bottomrule
\end{tabular}
}
\vspace{-5pt}
\caption*{(c) GPT-OSS-120B}
\end{minipage}

\\[1.2em] 

\begin{minipage}[t]{0.31\linewidth}
\centering\small
\resizebox{!}{0.31\linewidth}{
\begin{tabular}{lcccc}
\toprule
Metric & BS & MV & LFS & FFS \\
\midrule
Seq. tokens & 13.4k & 18.3k & 18.3k & \best{9.6k} \\
Total tokens & 107k & 106k & 106k & \best{10.4k} \\
\hdashline
GPQA & \best{68.2} & 66.7 & 61.6 & 66.2 \\
AIME24 & 76.7 & \best{83.3} & 76.7 & 80.0 \\
AIME25-I & 60.0 & \best{73.3} & 53.3 & 60.0 \\
AIME25-II & 60.0 & \best{86.7} & 66.7 & 80.0 \\
\bottomrule
\end{tabular}
}
\vspace{-5pt}
\caption*{(d) QwQ-32B}
\end{minipage}
\quad
\begin{minipage}[t]{0.31\linewidth}
\centering\small
\resizebox{!}{0.31\linewidth}{
\begin{tabular}{lcccc}
\toprule
Metric & BS & MV & LFS & FFS \\
\midrule
Seq. tokens & 12.4k & 19.3k & 19.3k & \best{3.5k} \\
Total tokens & 98.9k & 89.6k & 89.6k & \best{3.9k} \\
\hdashline
GPQA & \best{69.2} & \best{69.2} & 66.7 & 63.1 \\
AIME24 & 83.3 & \best{90.0} & 83.3 & 40.0 \\
AIME25-I & 66.7 & \best{73.3} & \best{73.3} & 40.0 \\
AIME25-II & 80.0 & \best{86.7} & 73.3 & 40.0 \\
\bottomrule
\end{tabular}
}
\vspace{-5pt}
\caption*{(e) Qwen3-32B}
\end{minipage}
\vspace{4pt}
\quad
\begin{minipage}[t]{0.31\linewidth}
\centering\small
\resizebox{!}{0.31\linewidth}{
\begin{tabular}{lcccc}
\toprule
Metric & BS & MV & LFS & FFS \\
\midrule
Seq. tokens & 4.1k & 5.7k & 5.7k & \best{3.8k} \\
Total tokens & 33.1k & 35.8k & 35.8k & \best{16.7k} \\
\hdashline
GPQA & 66.7 & 70.7 & 64.1 & \best{71.2} \\
AIME24 & 33.3 & \best{83.3} & \best{83.3} & \best{83.3} \\
AIME25-I & 40.0 & \best{53.3} & \best{53.3} & \best{53.3} \\
AIME25-II & 13.3 & \best{40.0} & \best{40.0} & \best{40.0} \\
\bottomrule
\end{tabular}
}
\vspace{-5pt}
\caption*{(f) Qwen3-235B}
\end{minipage}
\vspace{4pt}

\\[1.2em] 

\begin{minipage}[t]{0.31\linewidth}
\centering\small
\resizebox{!}{0.31\linewidth}{
\begin{tabular}{lcccc}
\toprule
Metric & BS & MV & LFS & FFS \\
\midrule
Seq. tokens & 8.7k & 14.6k & 14.6k & \best{6.4k} \\
Total tokens & 69.8k & 78.5k & 78.5k & \best{6.7k} \\
\hdashline
GPQA & 72.2 & \best{74.2} & 71.7 & 73.7 \\
AIME24 & 70.0 & 83.3 & 70.0 & \best{86.7} \\
AIME25-I & 60.0 & \best{73.3} & 53.3 & 66.7 \\
AIME25-II & 73.3 & \best{86.7} & 46.7 & 73.3 \\
\bottomrule
\end{tabular}
}
\vspace{-5pt}
\caption*{(g) R1}
\end{minipage}
\quad
\begin{minipage}[t]{0.31\linewidth}
\centering\small
\resizebox{!}{0.31\linewidth}{
\begin{tabular}{lcccc}
\toprule
Metric & BS & MV & LFS & FFS \\
\midrule
Seq. tokens & 12.2k & 17.9k & 17.9k & \best{6.0k} \\
Total tokens & 97.7k & 87.9k & 87.9k & \best{6.6k} \\
\hdashline
GPQA & 59.6 & \best{64.6} & 60.1 & 52.5 \\ 
AIME24 & 60.0 & 80.0 & 53.3 & \best{83.3} \\
AIME25-I & 46.7 & \best{60.0} & 46.7 & 53.3 \\
AIME25-II & \best{66.7} & 60.0 & 53.3 & 40.0 \\
\bottomrule
\end{tabular}
}
\vspace{-5pt}
\caption*{(h) R1-Distill-Qwen}
\end{minipage}
\end{tabular}
\caption{Accuracy (\%) and compute cost ($\times 10^{3}$ tokens) across models. For each method, token counts are dataset-averaged. Bold, gray cells mark the best value per row. Methods shown are beam search (BS), majority voting (MV), first finish search (FFS), last finish search (LFS).}
\end{table*}

\paragraph{Overall performance patterns.}
The four decoding strategies exhibit complementary trade-offs. MV is the most consistent accuracy-oriented method but incurs the largest token costs (often an order of magnitude higher than FFS). BS occasionally attains top accuracy on specific rows but is typically expensive. LFS shows mixed reliability: it matches or surpasses MV on some symbolic tasks (e.g., AIME variants for GPT-OSS-120B and Qwen3-235B) but underperforms on several models. FFS delivers substantial token savings (commonly reducing MV token usage by tens of percent up to $\sim$90\%), yet its accuracy impact is model-dependent—competitive in some cases (e.g., Dapo-Qwen-32B, Qwen3-235B, certain R1 entries) and substantially lower in others (e.g., DeepSeek, Qwen3).

\paragraph{Model-family comparison}
\begin{itemize}[leftmargin=4mm]
\item \textbf{Qwen-derived models (Qwen3-32B, Qwen3-235B-Instruct):} MV and LFS often lead on math tasks (AIME24/25), while FFS gives the largest efficiency gains but is not uniformly the accuracy winner. 
\item \textbf{DeepSeek family (DeepSeek, R1, R1-Distill-Qwen):} MV is the most stable high-accuracy choice; LFS sometimes rivals MV, and FFS reduces compute dramatically but with variable accuracy trade-offs.
\item \textbf{General-purpose large models (GPT-OSS-120B, QwQ-32B):} MV/LFS usually secure top accuracy; FFS offers strong efficiency with mixed accuracy effects (near-match in some cases, notable drops in others).
\end{itemize}

\paragraph{Task-level behavior.}
GPQA is relatively robust across decoding methods (small accuracy spread); FFS often preserves GPQA performance while saving tokens. AIME24 and the AIME25 subsets are more sensitive: MV and LFS usually perform better on structured symbolic reasoning, whereas FFS can be competitive for certain large models but may degrade accuracy for others.

\paragraph{Takeaway.}
There is no single best decoding method across models and tasks. MV remains the safest option for accuracy at higher compute cost; LFS is a viable middle ground for symbolic problems; BS is occasionally useful but expensive; FFS is attractive when compute is constrained but requires careful evaluation per model/task due to mixed accuracy outcomes.

\section{Model-wise Plots}
\label{appx:plots}

\begin{figure}[h]
    \centering
    \includegraphics[width=\linewidth]{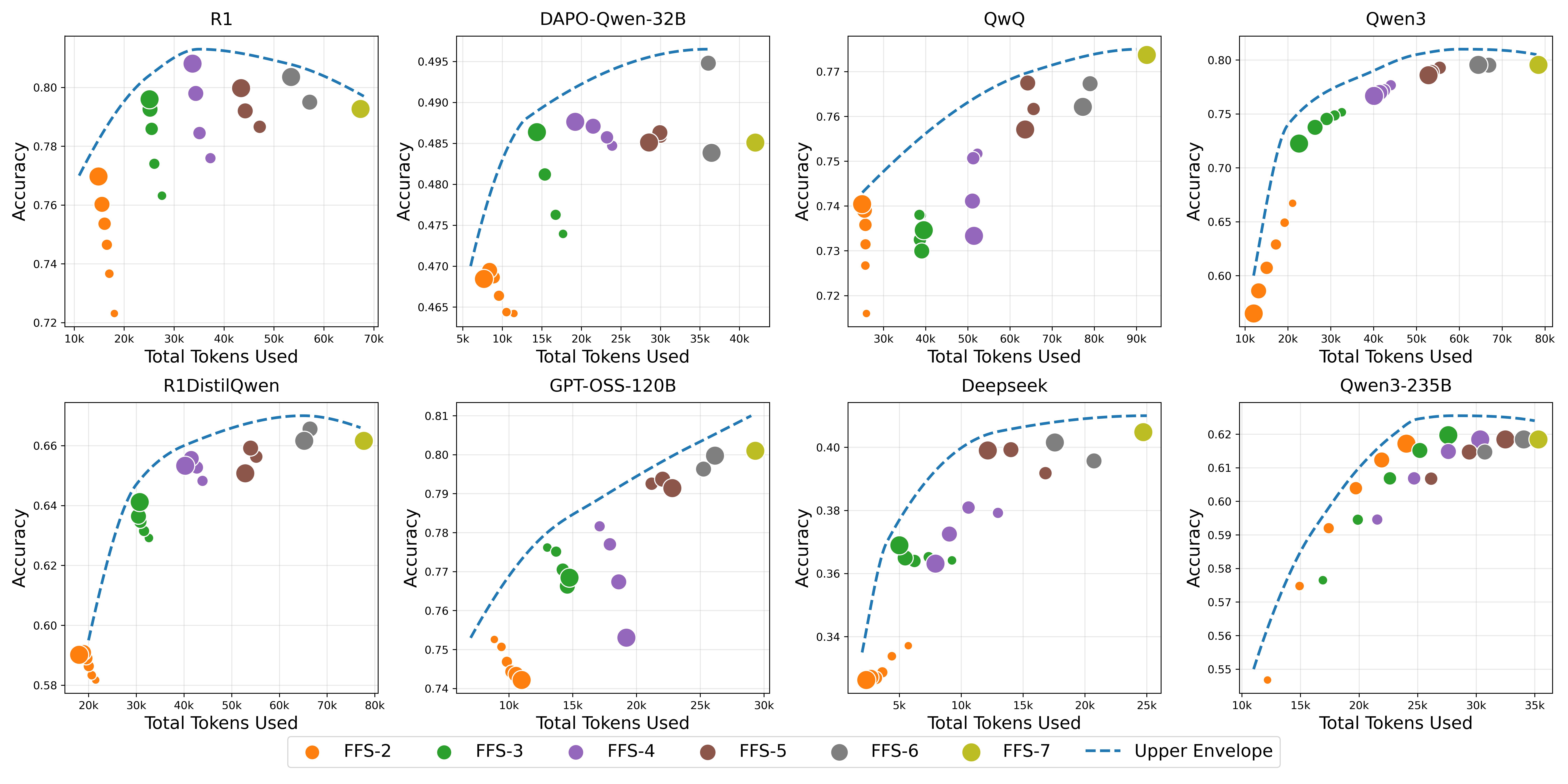}
    \caption{Accuracy versus token usage for different models. FFS-k variants are
shown in distinct colors (one color per k). Marker size encodes the value of N, with larger
markers representing larger N.}
    \label{fig:ffs-all-models}
\end{figure}

\begin{figure}[h]
    \centering
    \includegraphics[width=\linewidth]{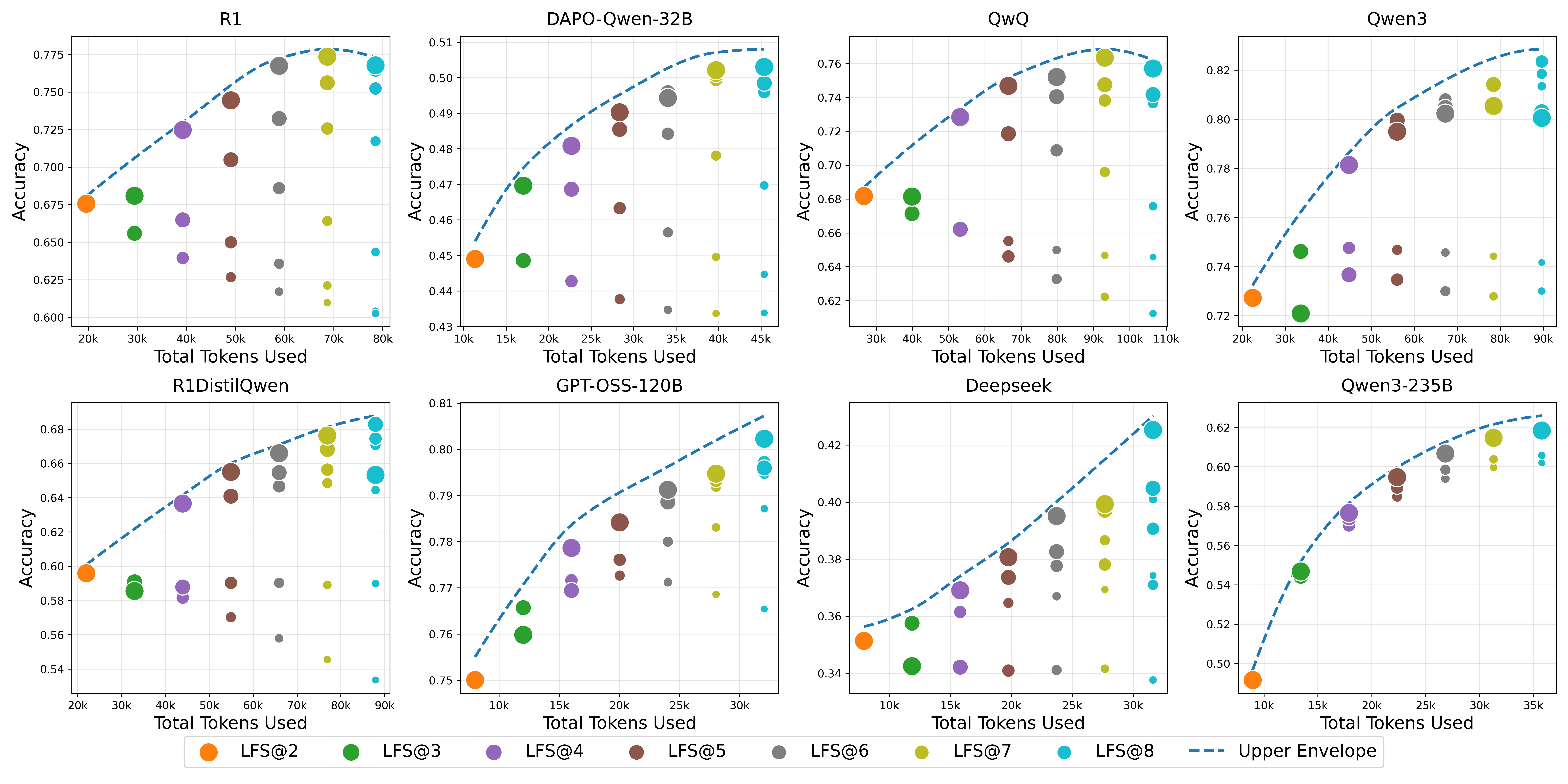}
    \caption{Accuracy versus token usage for different models. LFS@N variants are shown in distinct colors (one color per N). Marker size encodes the value of k, with larger markers representing larger k.}
    \label{fig:lfs-all-models}
\end{figure}

Figures~\ref{fig:ffs-all-models} and \ref{fig:lfs-all-models} contain the FFS and LFS curves, for each of the eight models individually. 

